%% file: main.tex
\documentclass{article}
\usepackage[utf8]{inputenc}
\usepackage{graphicx}
\usepackage{microtype}
\usepackage{float}
\usepackage{subfigure}
\usepackage{booktabs} 
\usepackage{amsmath}
\usepackage{amssymb}
\usepackage{tikz}
\usepackage{comment}
\usetikzlibrary{arrows,decorations.markings}

\usepackage{algorithm}
\usepackage[noend]{algorithmic}

\usepackage{hyperref}

\usepackage[accepted]{icml2018}

\icmltitlerunning{Improving width-based planning with compact policies
}

\begin{document}

\twocolumn[
\icmltitle{Improving width-based planning with compact policies}



\icmlsetsymbol{equal}{*}

\begin{icmlauthorlist}
\icmlauthor{Miquel Junyent}{oesia,upf}
\icmlauthor{Anders Jonsson}{upf}
\icmlauthor{Vicenç Gómez}{upf}
\end{icmlauthorlist}

\icmlaffiliation{oesia}{Grupo Oesía, Barcelona, Spain}
\icmlaffiliation{upf}{Universitat Pompeu Fabra, Barcelona, Spain}

\icmlcorrespondingauthor{}{mjunyent@oesia.com}

\icmlkeywords{Machine Learning, ICML}

\vskip 0.3in
]



\printAffiliationsAndNotice{\noindent \textbf{ICML 2018 workshop: Planning and Learning}.\\}  

\begin{abstract}

Optimal action selection in decision problems characterized by sparse, delayed rewards is still an open challenge. For these problems, current deep reinforcement learning methods require enormous amounts of data to learn controllers that reach human-level performance. In this work, we propose a method that interleaves planning and learning to address this issue. The planning step hinges on the Iterated-Width (IW) planner, a state of the art planner that makes explicit use of the state representation to perform structured exploration. IW is able to scale up to problems independently of the size of the state space. From the state-actions visited by IW, the learning step estimates a compact policy, which in turn is used to guide the planning step. The type of exploration used by our method is radically different than the standard random exploration used in RL. We evaluate our method in simple problems where we show it to have superior performance than the state-of-the-art reinforcement learning algorithms A2C and Alpha Zero. Finally, we present preliminary results in a subset of the Atari games suite.

\end{abstract}

\input{contents/intro}

\input{contents/background}

\input{contents/related}

\input{contents/approach}

\input{contents/experiments}

\input{contents/conclusions}

\section*{Acknowledgements}
Miquel Junyent's research is partially funded by project 2016DI004 of the Catalan Industrial Doctorates Plan. Anders Jonsson is partially supported by the grants TIN2015-67959 and PCIN-2017-082 of the Spanish Ministry of Science.
Vicen\c{c} G\'omez is supported by the Ramon y Cajal program RYC-2015-18878 (AEI/MINEICO/FSE,UE).

\bibliographystyle{icml2018}
\bibliography{bibliography}

\end{document}

%% file: contents/intro.tex
\section{Introduction}\label{sec:intro}
Optimal sequential decision making is a fundamental problem to many diverse fields.
In the recent years the Reinforcement Learning (RL) approach has experienced unprecedented success, reaching human-level performance in several domains, including Atari video-games~\cite{Mnih2015Human-levelLearning} or the ancient game of Go~\cite{silver2016mastering}. 
This success has been largely enabled by the use of advanced function approximation techniques in combination with large-scale data generation from self-play games. Current RL methods, however, still require enormous amounts of data to learn,
specially in tasks characterized by delayed, sparse rewards,  
where more efficient ways of exploring the problem state space are needed.


Safe online exploration can be incentivized by adding a reward bonus. 
This is known under different names: reward shaping~\cite{ng1999policy}, optimism in the face of uncertainty~\cite{kearns2002near}, intrinsic motivation~\cite{NIPS2004_2552}, curiosity-driven RL~\cite{still2012information}, prediction gain~\cite{Bellemare2016UnifyingMotivation}, or entropy-regularized MDPs~\cite{neu2017unified}.
Alternative approaches introduce correlated noise directly in the parameter space of the policy or value function~\cite{plappert2018parameter,fortunato2018noisy,Osband2016DeepDQN,liu2017stein}.

While these approaches offer significant improvements over classical exploration techniques such as $\epsilon$-greedy or Boltzmann exploration, none of them 
makes explicit use of the representation of the state, which is treated as a black box.
The traditional way to exploit the state structure in RL is through the framework of factored MDPs.
Factored MDPs represent compactly the transition probabilities of the MDP in terms of products of factors that involve a small subset of the state variables, allowing to reduce exponentially the sample complexity in some cases~\cite{boutilier2000stochastic,kearns1999efficient}. 
However, the factorization structure is in general not inherited by the value function and is not generally exploited to guide the exploration.


In contrast to the RL approach in which the agent learns a policy by interacting with the environment, the planning approach for decision making assumes known models for the agent's goals and domain dynamics, and focuses on determining how the agent should behave to achieve its objectives~\cite{kolobov2012planning}. 
Current planners are able to solve problem instances involving huge state spaces by precisely exploiting the problem structure that is defined in the state-action model~\cite{Planning}. 

A family of planners, known as width-based planners, was introduced by \citet{Lipovetzky2012WidthProblems} and became the state-of-the-art for solving planning benchmarks. While originally proposed for classical planning problems, i.e., deterministic, goal-driven problems with a fully defined model, width-based planners have evolved closer to the RL setting.
Recently, these planners have been applied to Atari games using pixel features reaching comparable results with learning methods in almost real-time~\cite{bandres2018planning}.


In this paper, we explore further this family of methods originated in the planning community. 
In particular, we consider width-based planning in combination with a learning policy, showing that this combination has benefits both in terms of exploration and feature learning. 
Our approach is to train a policy in the form of a neural network, and use
the compact state representation learned by the policy  
to guide the width-based planning algorithm. We show that the lookahead power of the resulting algorithm is comparable to or better than previous width-based implementations that use static features. 
Our approach is based on the same principle as the recently proposed AlphaZero~\cite{Silver2017MasteringChess}, which also interleaves learning with planning, but uses a Monte-Carlo tree search planner instead.

The next section introduces the basic background and Section~\ref{sec:rw} describes related work. We then present our approach in Section~\ref{sec:approach}, followed by experimental results in Section~\ref{sec:exp}. We conclude and outline future work directions in Section~\ref{sec:fin}.


%% file: contents/background.tex
\section{Background}

In this section, we review the fundamental concepts of reinforcement learning and the width-based planning.

\subsection{Reinforcement Learning and MDPs}

We consider sequential decision problems modelled as Markov decision processes (MDPs). An MDP is a tuple $M = \langle S,A,P,r \rangle$, where $S$ is the finite state space, $A$ is the finite action 
space, $P:S \times A \rightarrow \Delta(S)$ is the transition function,
and $r:S \times A \rightarrow \mathbb{R}$ is the reward function. 
Here, $\Delta(S)=\{\mu\in\mathbb{R}^{S}: \sum_s\mu(s)=1, \mu(s)\geq 0 \; (\forall s)\}$ is the probability simplex over $S$.

In each time-step $t$, the learner observes state $s_t \in S$, selects action $a_t \in A$, moves to the next state $s_{t+1} \sim P(\cdot|s_t,a_t)$, and obtains reward $r_{t+1}$ such that $\mathbb{E}[r_{t+1}]=r(s_t,a_t)$. We refer to the tuple $\langle s_t,a_t,r_{t+1},s_{t+1}\rangle$ as a {\em transition}. The aim of the learner is to select actions to maximize the expected cumulative discounted reward $\mathbb{E}\left[\sum_{k=t}^\infty \gamma^{k-t} r_{k+1}\right]$, where $\gamma\in(0,1]$ is a discount factor. We assume that the transition function $P$ and reward function $r$ are unknown to the learner.

The decision strategy of the learner is represented by a {\em policy} $\pi:S \rightarrow \Delta(A)$, i.e.~a mapping from states to distributions over actions. A policy~$\pi$~induces a {\em value function} $V^\pi:S\rightarrow\mathbb{R}$ such that for each state $s$, $V^\pi(s)=\mathbb{E}\left[\sum_{k=t}^\infty \gamma^{k-t} r_{k+1} | \, s_t=s\right]$ is the expected cumulative reward when starting in state $s$ and using policy $\pi$ to select actions. The {\em optimal value function} $V^*$ achieves the maximum value in each state $s$, i.e.~$V^*(s)=\max_\pi V^\pi(s)$, and the {\em optimal policy} $\pi^*$ is the policy that attains this maximum in each state $s$, i.e.~$\pi^*(\cdot|s)=\arg\max_\pi V^\pi(s)$. Typically, an estimate $\widehat\pi_\theta$ of the optimal policy and/or an estimate $\widehat V_\varphi$ of the optimal value function are maintained, parameterized on vectors $\theta$ and $\varphi$, respectively.

\subsection{Width-based Planning}

Iterated Width (IW) is a pure exploration algorithm originally developed for goal-directed planning problems with deterministic actions~\cite{Lipovetzky2012WidthProblems}. It requires the state space to be factored into a set of {\em features} or {\em atoms} $\Phi$. We assume that each feature has the same domain $D$, e.g.~binary ($D=\{0,1\}$) or real-valued ($D=\mathbb{R}$). The algorithm consists of a sequence of calls IW($i$) for $i=0,1,2,\hdots$ until a termination condition is reached.
IW($i$) performs a standard breadth-first search from a given initial state~$s_0$, but prunes states that are not {\em novel}.
When a new state $s$ is generated, IW($i$) contemplates all $n$-tuples of atoms of $s$ with size $n \leq i$. The state is considered novel if at least one tuple has not appeared in the search before, otherwise it is pruned.

IW($i$) is thus a $\emph{blind}$ search algorithm that traverses the entire state-space for sufficiently large $i$. The traversal is solely determined by how the state is structured, i.e., what features are used to represent the state.
Each iteration IW($i$) is an $i$-width search that is complete for problems whose width is bounded by $i$ and has complexity $\mathcal{O}(|\Phi|^i)$, where $|\Phi|$ is the number of problem variables or features~\cite{Lipovetzky2012WidthProblems}.
Interestingly, most planning benchmarks turn out to have very small width and, in practice, they can be solved in linear or quadratic time.




\begin{figure}[t]
\begin{center}
\begin{tikzpicture}[scale=0.70]
    \node (a) at (3,2) [draw] {$(0,0,0)$};
    \node (b) at (0,0) [draw] {$(1,0,1)$};
    \node (c) at (2,0) [draw] {$(1,0,0)$};
    \node (d) at (4,0) [draw] {$(0,1,1)$};
    \node (e) at (6,0) [draw] {$(0,0,1)$};

    \draw [decoration={markings,mark=at position 1 with {\arrow[scale=2,>=stealth]{>}}},postaction={decorate}] (a) -- (b);
    \draw [decoration={markings,mark=at position 1 with {\arrow[scale=2,>=stealth]{>}}},postaction={decorate}] (a) -- (c);
    \draw [decoration={markings,mark=at position 1 with {\arrow[scale=2,>=stealth]{>}}},postaction={decorate}] (a) -- (d);
    \draw [decoration={markings,mark=at position 1 with {\arrow[scale=2,>=stealth]{>}}},postaction={decorate}] (a) -- (e);

    \draw [color=red] (1.2,-0.4) -- (2.8,0.4);
    \draw [color=red] (1.2,0.4) -- (2.8,-0.4);
    \draw [color=red] (5.2,-0.4) -- (6.8,0.4);
    \draw [color=red] (5.2,0.4) -- (6.8,-0.4);
\end{tikzpicture}
\end{center}
\caption{Example run of IW(1). States are represented by their feature vectors, and actions correspond to edges.}
\label{fig:iw}
\end{figure}
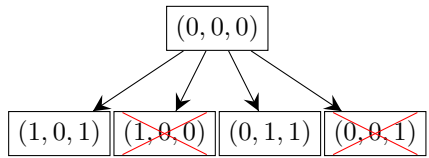

We illustrate IW(1) using a small example that involves three binary features (i.e.~$D=\{0,1\}$) and four actions. Figure~\ref{fig:iw} shows an example run of IW(1), in which the initial state $s_0$ maps to the feature vector $(0,0,0)$. Assume that breadth-first search expands the states at depth $1$ in the order left-to-right. The first state generates two new feature values: the first and third feature have value $1$, and therefore is not pruned. 
The second state does not generate any new feature values, and is thus pruned. The third state assigns $1$ to the second feature for the first time, while the fourth state is also pruned.
The algorithm continues expanding the nodes that have not been pruned in a breath-first manner until all nodes are pruned or the goal state is reached.

%% file: contents/related.tex
\section{Related work}\label{sec:rw}

In the following subsections, we review relevant literature from three different viewpoints: possible extensions to the original IW algorithm, efficient exploration in RL, and recently proposed methods that combine planning and RL.

\subsection{Width-Based planning for MDPs}

There are two follow-ups to the original IW algorithm of special relevance to this work. First, \citet{Lipovetzky2015ClassicalGames} extended the original algorithm to MDPs by associating a reward $R(s)$ to each state $s$ during search, equivalent to the reward $\sum_{t=0}^{d-1} \gamma^t r_{t+1}$ accumulated on the path from $s_0$ to $s_d=s$, where $d$ is the depth of $s$ in the search tree.
The discount factor $\gamma$ has the effect of favoring earlier rewards.
With this extension, after the search completes, the first action on the path from $s_0$ to $s^*$ is applied, similarly to model predictive control. This version of IW(1) achieves competitive performance in the Atari suite~\cite{Bellemare2012TheAgents}, using as features the 128 bytes of the RAM memory representing the current game configuration.

\citet{bandres2018planning} further modified the algorithm to handle visual features. Specifically, their algorithm uses the (binary) B-PROST features extracted from the images of Atari games~\cite{Liang2016StateLearning}. Since storing and retrieving Atari states during breadth-first search is costly, the authors introduced a Rollout version of IW(1) which \emph{emulates} the breadth-first traversal of the original algorithm, by keeping track of the minimum depth at which a feature is found for the first time and extending the notion of novelty accordingly. The pruned states are kept as leaves of the tree, and are considered as candidates for states with highest reward.

The above contributions brought the original formulation of IW closer to the RL setting. 
The rollout version of IW only requires a simulator (a successor function from a given state) and a structured representation of the state in terms of atoms or features.  
However, two important challenges remain. First, IW is used in an open-loop way and does not produce a compact policy that can be used in a reactive, closed-loop environment. Second, width-based algorithms use a fixed set of features that needs to be defined in advance. While competitive performance has been achieved using pixels as features in the Atari domain, interesting states may still require a large width to be reached, which can be unfeasible. These two challenges are the main motivation for our work.


\subsection{Exploration in Reinforcement Learning}


As mentioned in the introduction, there are several alternative approaches to efficient exploration in RL. Most of these approaches are based on the idea of adding an explicit bonus to the reward function.
The intuition is that by adding a bonus to states that have not been frequently visited during search, the likelihood for visiting unexplored parts of the state space will increase, potentially leading to higher rewards. Even though this scheme does not preserve the Markov property (since the reward now typically depends on the number of times we have visited a state), the result is often that the learner ends up exploring larger sections of the state space. This follows the well-known principle of optimism in the face of uncertainty~\cite{kearns2002near}.


The UCT algorithm for Monte-Carlo tree search~\cite{UCT}, and its precursor UCB for multi-armed bandits~\cite{UCB}, are examples of algorithms that assign a reward bonus to states which is inversely proportional to the number of times that a state has been visited during search. Since the search tree has finite size, it is feasible to count the number of times each state is visited.

When the state space is very large, maintaining an explicit visitation count becomes infeasible.  \citet{Bellemare2016UnifyingMotivation} address this problem by introducing {\em pseudo-counts} based on the {\em prediction gain} of states. The authors show that the pseudo-counts are good approximators of how often states are visited, and define a reward bonus inversely proportional to the pseudo-count. \citet{Martin2017Count-BasedLearning} further extend the idea of pseudo-counts from states to feature vectors. The idea is to decompose computation of the pseudo-count such that the pseudo-count of a state $s$ is composed of the pseudo-counts of each feature of $s$. This significantly simplifies the computation and reduces the algorithmic complexity.

Compared to the above approaches, IW(1) is a pure exploration algorithm that does not take reward into account at all. The only purpose of tracking the reward of states is to decide which action to perform when the search concludes.

\subsection{Combining Reinforcement Learning and Planning} 

A natural way to combine planning and learning is to identify the planner as a ``teacher" that provides \emph{correct} transitions that are used to learn a policy,
as in imitation learning~\cite{DAGGER,guo2014deep}.
Recently, AlphaGo achieved superhuman performance in the game of Go \cite{silver2016mastering} by combining supervised learning from expert moves and self-play. 
AlphaZero~\cite{Silver2017MasteringKnowledge}, a version of the same algorithm that learned solely from self-play, has outperformed previous variants, also showing stunning results in Chess and Shogui~\cite{Silver2017MasteringChess}.



At every iteration $t$, AlphaZero generates a tree using Monte-Carlo tree search, guided by a policy and a value estimate. It keeps a visit count on the branches of the tree, and uses it to explore less frequent states (using a variant of UCT) and to generate a target policy $\pi^{target}_t$. After tree expansion, an action is selected at the root following $\pi^{target}_t$, and the resulting subtree is kept for the next iteration. At the end of the episode, the win/loss result $z_t$ is recorded and all transitions $(s_t, \pi^{target}_t, z_t)$ are added to a dataset. In parallel, the policy and value estimates are trained in a supervised manner with minibatches sampled from the dataset.  


%% file: contents/approach.tex
\section{Policy-guided Iterated Width}\label{sec:approach}
We now present our proposed algorithm to combine planning and learning. Our aim is two-fold. On the one hand, in spite of its success, IW does not learn from experience, so its performance does not improve over time. On the other hand, RL algorithms usually suffer from poor exploration, struggling to solve problems with sparse rewards and significantly slowing down learning.

In this work we leverage the exploration capacity of IW(1) to train a policy estimate $\widehat\pi_\theta$.  
Both IW and Rollout IW select actions at random with uniform probabilities. Even though both algorithms favor novel states with previously unseen feature values, random action selection does not take into account previous experience and results in a uninformed exploration. As a result, reaching a distant reward in a specific search may be arbitrary.
We build on the recently proposed Rollout based IW version \cite{bandres2018planning} by incorporating an action selection policy,
resulting in an \emph{informed} IW search.
The combination of IW and RL addresses the shortcomings of each approach, resulting in an efficient algorithm in terms of both exploration and learning a policy $\widehat\pi_\theta$ that can be used in closed-loop scenarios.

Our extension, Policy-guided Iterated Width (PIW), enhances Rollout IW by guiding the search with the current policy estimate $\widehat\pi_\theta$. We consider tuples of size $1$, i.e., IW(1), which keeps the planning step tractable. 
Similar to Rollout IW, PIW requires a simulator that provides the successor of a state $s$ and a representation of $s$ as features $\Phi$.

The algorithm interleaves a Rollout IW planning step with a policy learning step, which we describe next. 
After describing the basic PIW algorithm, we present a possible way to discover features using the learned policy that can in turn be used by IW.
This second use of the policy can be beneficial if the original features are poor or unknown.

\subsection{Planning step}

At every iteration, the algorithm first selects a node $n$ for expansion, and then performs a rollout from $n$. To find $n$, PIW uses the current policy $\widehat\pi_\theta$ to select actions. The tree is traversed until a state-action pair $(n, a)$ that has not yet been expanded is found. The rollout from $n$ also uses $\widehat\pi_\theta$ to select actions until a terminal state or a state that is not novel is reached. At that point, the final node is marked as \textit{solved} and the process restarts until all nodes have been solved or a maximum budget of time or nodes is exhausted. Algorithm \ref{algo} shows the planning step of PIW.

In this work, our policy takes the form of a neural network (NN) $\widehat\pi_\theta$ with softmax outputs
$\widehat\pi_\theta(a|s_n) = \frac{e^{h_a(s_n)/\tau}}{\sum_{b \in A}{e^{h_b(s_n)/\tau}}}$, 
where $h_a$, $a\in A$, are the logits output of the NN and $\tau$ is a temperature parameter for additional control of the exploration. 
We leverage finding a good representation of the state on the NN, which will be learned using samples from IW.

\input{contents/algorithm}

In the limit $\tau \rightarrow \infty$ we obtain the uniform policy used in width-based algorithms. Just as in Rollout IW, actions that lead to nodes that are labelled as solved should not be considered. Thus, we set probability $\widehat\pi_\theta(a|s_n) = 0$ for each solved action $a$ and normalize $\widehat\pi_\theta$ over the remaining actions before sampling (\textit{Select\_action\_following\_policy}).

Every time a node is labelled as solved, the label is propagated along the branch to the root. A node is labelled as solved if each of its children is labelled as solved (\textit{Solve\_and\_propagate\_label}). Initially, all nodes of the cached tree are marked as not solved, except for the ones that are terminal (\textit{Initialize\_labels}).

As previously mentioned, an expanded state is considered novel if one of its atoms is true at a smaller depth than the one registered so far in the novelty table. A node that was already in the tree will not be pruned if its depth is exactly equal to the one in the novelty table for one of its atoms.

\subsection{Learning step}

Once the tree has been generated, the discounted rewards are backpropagated to the root: $R_i = r_i + \gamma \max_{j \in children(i)}{R_j}$. A target policy $\pi^{target}_t(\cdot|s_t)$ is induced from the returns at the root node by applying a softmax with $\tau \rightarrow 0$, i.e. a one-hot encoding of the maximum return, except for the cases where more than one path leads to the same return, in which case each path is assigned equal probability. The state $s_t$ is stored together with the target policy to train the model in a supervised manner. 
We use the cross-entropy error between the induced target policy $\pi_{t}^{target}(\cdot|s_t)$ and the current policy estimate $\widehat\pi_\theta(\cdot|s_t)$ to update the policy parameters $\theta$, defining a loss function
\begin{align*}
l=-\pi_{t}^{target}(\cdot|s_t)^\top\log\widehat\pi_\theta(\cdot|s_t).
\end{align*}
L2 regularization may be added to avoid overfitting and help convergence.
The model is trained by randomly sampling transitions from the dataset, which can be done in parallel, as in AlphaZero, or a training step can be taken after each lookahead.
In our experiments we choose the latter, sampling a batch of transitions at each iteration.
We keep a maximum of $T$ transitions, discarding outdated transitions in a FIFO manner.

Finally, a new root is selected from the nodes at depth 1 following $a_t \sim \pi^{target}_t(\cdot|s_t)$ and the resulting subtree is kept for the next planning step. This has been referred to as \textit{tree caching} in previous work~\cite{Lipovetzky2012WidthProblems}, and it has been argued that not including cached nodes in the novelty table increases exploration and hence performance. Note that cached nodes will contain outdated information. Although we did not find this to have a great impact on performance, one possibility is to rerun the model on all nodes of the tree at regular intervals. However, this is not done in our experiments.

\subsection{Dynamic features}

The quality of the transitions recorded by IW greatly depends on the feature set $\Phi$ used to define the novelty of states. For example, even though IW has been applied directly to visual (pixel) features~\cite{bandres2018planning}, it tends to work best when the features are {\em symbolic}, e.g., when the RAM state is used as a feature vector~\cite{Lipovetzky2015ClassicalGames}.
Symbolic features makes planning become more effective, since the width of a problem is effectively reduced by the information encoded in the features.
However, how to automatically learn powerful features for this type of structured exploration is an open challenge.

In PIW, we can use the representation learned by the NN to define a feature space, as in representation learning~\cite{Goodfellow-et-al-2016}.
With this dependence, the behavior of IW effectively changes when interleaving policy updates with runs of IW. If appropriately defined, these features should help to distinguish between important parts of the state space. In this work, we extract $\Phi$ from the last hidden layer of the neural network. In particular, we use the output of the rectified linear units that we subsequently discretize in the simplest way, resulting in binary features ($0$ for negative outputs and $1$ for positive outputs).

%% file: contents/algorithm.tex
\begin{algorithm}[!t]
\caption{Planning step of Policy-Guided IW(1)}
\label{algo}
\begin{algorithmic}
	\FUNCTION{Generate\_lookahead\_tree(tree)}
    \STATE Initialize\_labels(tree)
	\STATE D := Make\_empty\_novelty\_table()
 	\WHILE{within\_budget \AND $\neg$tree.root.solved}
    	\STATE n, a := Select(tree.root, D)
        \IF{a $\neq \; \perp$}
			\STATE Rollout(n, a, D)
        \ENDIF
    \ENDWHILE
    \ENDFUNCTION
    
    \STATE 
    
    \FUNCTION{Select(n, D)}
    \LOOP
    	\STATE novel := Check\_novelty(D, n.atoms, n.depth, \textbf{false})
    	\IF{is\_terminal(n) \OR $\neg$novel}
        	\STATE Solve\_and\_propagate\_label(n)
        	\STATE \textbf{return} n, $\perp$
    	\ENDIF
    	\STATE a := Select\_action\_following\_policy(n)
    	\IF{n[a] \textbf{in} tree}
    		\STATE n := n[a]
    	\ELSE
    		\STATE \textbf{return} n, a
    	\ENDIF
    \ENDLOOP
    \ENDFUNCTION
    
    \STATE 
    
    \FUNCTION{Rollout(n, a, D)}
    \WHILE{within\_budget}
		\STATE n := expand\_node(n, a)
		\STATE n.solved := \textbf{false}

		\STATE novel := Check\_novelty(D, n.atoms, n.depth, \textbf{true})
    	\IF{is\_terminal(n) \OR $\neg$novel}
    		\STATE Solve\_and\_propagate\_label(n)
        	\STATE \textbf{return}
    	\ENDIF
        \STATE a := Select\_action\_following\_policy(n)
    \ENDWHILE
    \ENDFUNCTION
    
    \STATE 
    
    \FUNCTION{Check\_novelty(D, atoms, d, is\_new)}
	\STATE novel := \textbf{false}
	\FOR{f \textbf{in} atoms}
    	\STATE novel := novel $\lor$ d $<$ D[f] $\lor$ ($\neg$is\_new $\land$ d = D[f])
    	\IF{d $<$ D[f] $\land$ is\_new}
            \STATE D[f] := d
        \ENDIF
    \ENDFOR
    \STATE \textbf{return} novel
    \ENDFUNCTION
\end{algorithmic}
\end{algorithm}

%% file: contents/experiments.tex
\section{Experiments}\label{sec:exp}

We evaluate the performance of Policy-guided IW in different settings. First, we consider a toy problem where we compare our method against state-of-the-art RL algorithms. We show that PIW is superior to current methods in a challenging, sparse-reward environment. Second, we present preliminary results in large-scale problems, testing our approach in four Atari 2600 games, and we show that PIW outperforms previous width-based approaches. Finally, we compare the policy learned by PIW with state-of-the-art results in the Atari benchmark.


\subsection{Simple environments}

\begin{figure}
\centering
\includegraphics[width=0.25\columnwidth]{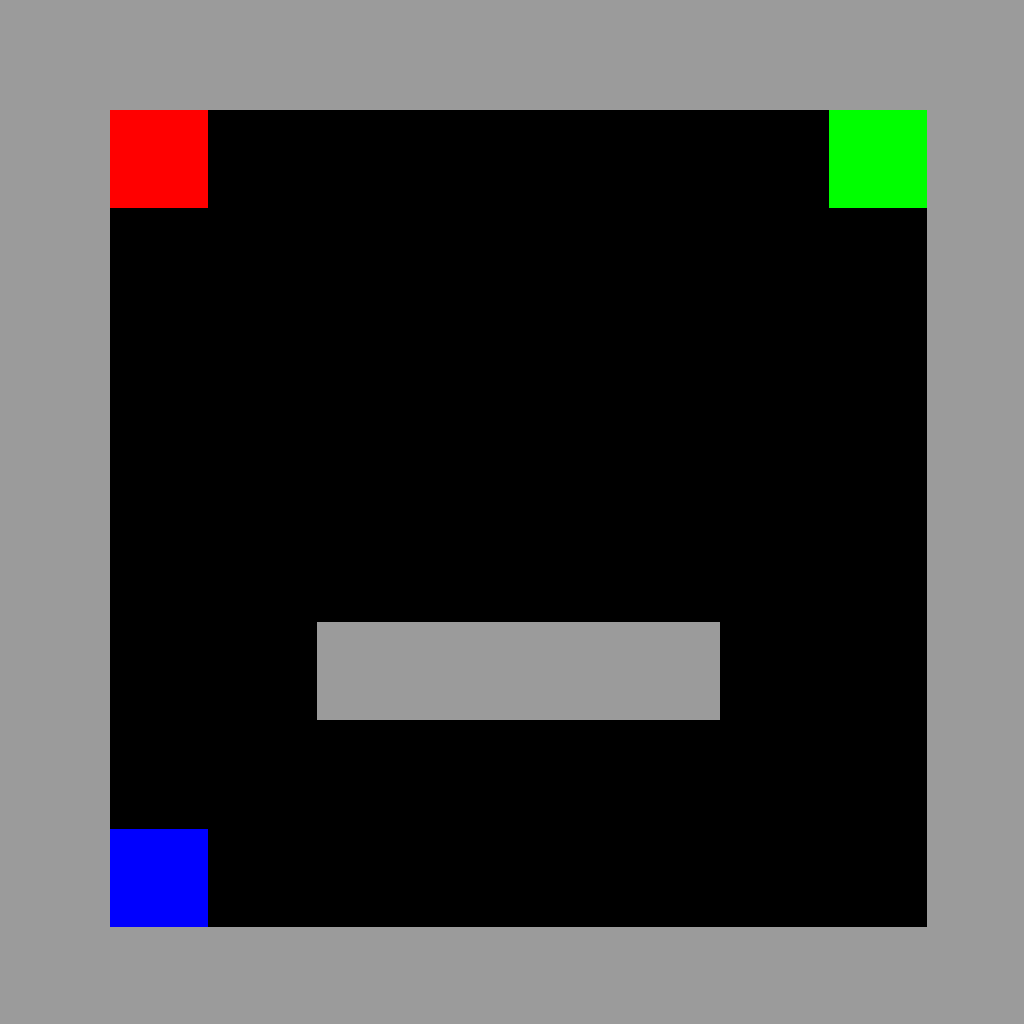}
\hspace{0.05\columnwidth}
\includegraphics[width=0.25\columnwidth]{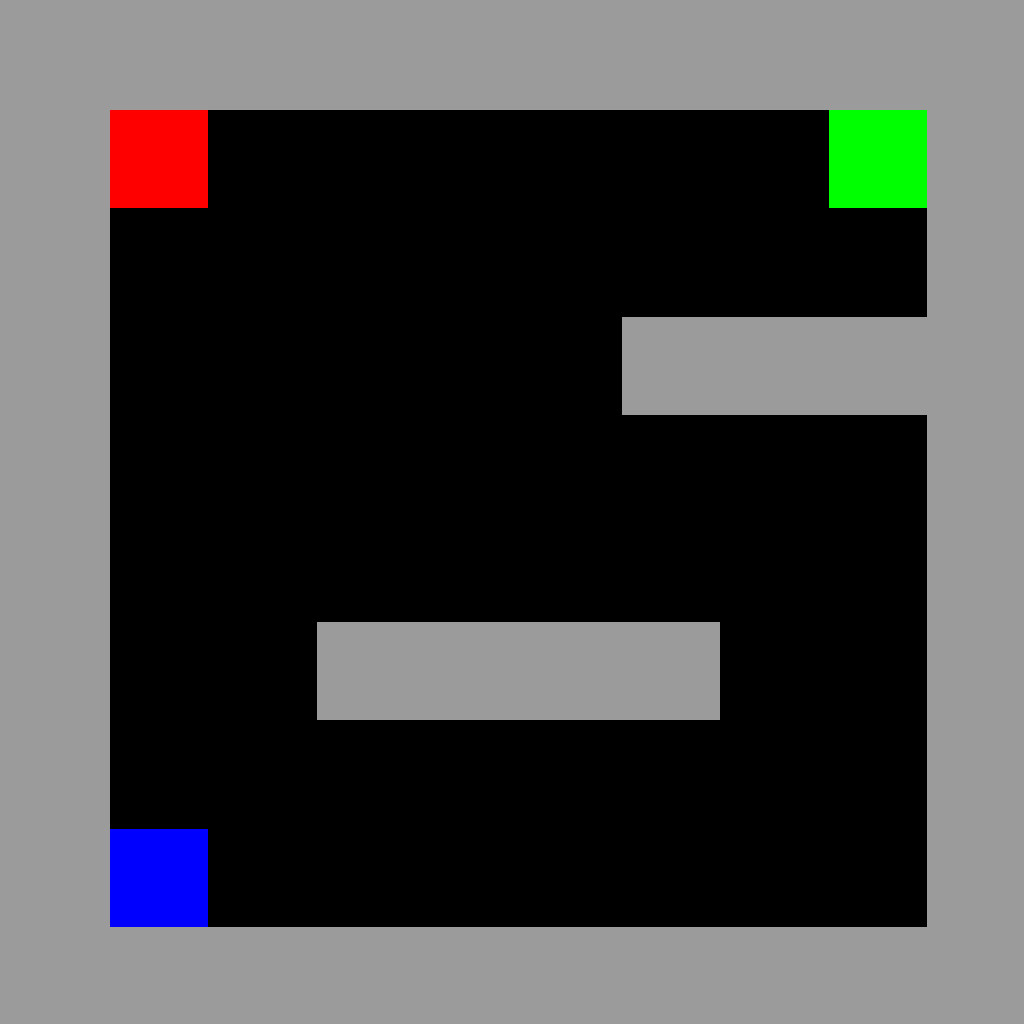}
\hspace{0.05\columnwidth}
\includegraphics[width=0.25\columnwidth]{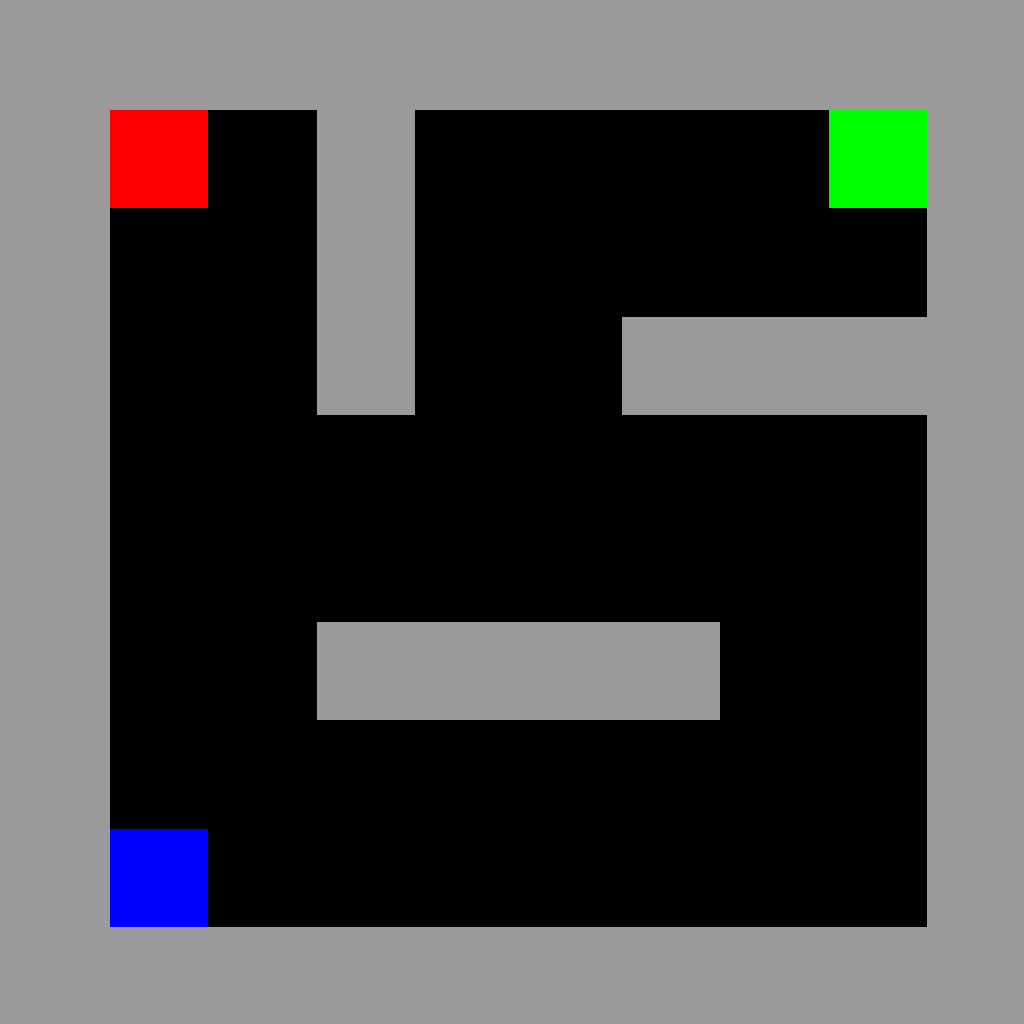}
\caption{Snapshot of the three versions of the game. The blue, green and red squares represent the agent, the key and the door respectively.}
\label{fig:gridworld}
\end{figure}

To test our approach, we use a $10 \times 10$ gridworld environment where an agent has to navigate to first pick up a key and then go through a door. An episode terminates with a reward of $+1$ when the goal is accomplished, with a reward of $-1$ when a wall is hit or with no reward after a maximum of 200 steps is reached. All intermediate states are not rewarded. The observation is an $84 \times 84$ RGB image and possible actions are no-op, going up, down, left or right.
The environment is challenging since the reward is sparse and each episode terminates when the agent hits a wall (resetting the agent's position). We consider three variants of the game, with increasing difficulty (see Figure \ref{fig:gridworld}).

We compare our approach to AlphaZero. Although originally designed for two-player zero-sum games, it can be easily extended to the MDP setting. AlphaZero controls the balance between exploration and exploitation by a parameter $p_{uct}$ together with a temperature parameter in the target policy $\tau$, similar to ours. In the original paper, $\tau$ is set to 1 for a few steps at the beginning of every episode, and then it is changed to an infinitesimal temperature $\tau=\epsilon$ for the rest of the game \cite{Silver2017MasteringKnowledge}. Nevertheless, we achieved better results in our experiments with AlphaZero using $\tau = 1$ for the entire episode. Furthermore, AlphaZero needs to wait until the episode ends to assign a target value for all transitions in the episode. Thus, for a fair comparison, in these experiments, PIW also adds all the transitions of an episode to the dataset upon termination.

We analyze PIW using static and dynamic features. For the first case, we take the set of BASIC features \cite{Bellemare2012TheAgents}, where the input image is divided in tiles and an atom, represented by a tuple $(i,j,k)$, is true if color $k$ appears in the tile $(i,j)$. In our simple environment, we make the tiles coincide with the grid, and since there is only one color per tile, the amount of features is limited to 100. For the second case, we take the (discretized) outputs of the last hidden layer as binary feature vectors.

All algorithms share the same NN architecture and hyperparameters, specified in Table \ref{tab:hyperparameters}. We use two convolutional and two fully connected layers as in \citet{mnih2013playing}, and we train it using the non-centered version of the RMSProp algorithm. Although all algorithms can be run in parallel, the experiments presented in this paper have been executed using one thread.

\begin{table}
\caption{Hyperparameters used for PIW, AlphaZero and A2C.}
\label{tab:hyperparameters}
\begin{center}
\begin{small}
\begin{tabular}{lcc}
\toprule
Hyperparameter &  Value & Algorithm\\
\midrule
Discount factor & 0.99 & All \\
Batch size & 10 & All \\
Learning rate &  0.0007 & All \\
Clip gradient norm & 40 & All \\
RMSProp decay & 0.99 & All \\
RMSProp epsilon & 0.1 & All \\
Tree budget nodes & 50 & PIW, AlphaZero \\
Dataset size $T$ & $10^3$ & PIW, AlphaZero \\
L2 reg. loss factor & $10^{-4}$ & PIW, AlphaZero \\
Tree policy temp. $\tau$ & 1 & PIW, AlphaZero \\
$p_{uct}$ & 0.5 & AlphaZero\\
Diritchlet noise $\alpha$ & 0.3 & AlphaZero \\
Noise factor & 0.25 & AlphaZero \\
Value loss factor & 1 & AlphaZero, A2C \\
Entropy loss factor & 0.1 & A2C \\
\bottomrule
\end{tabular}
\end{small}
\end{center}
\vskip -0.1in
\end{table}

\begin{figure*}[t]
\centering

\includegraphics[height=0.179\textheight]{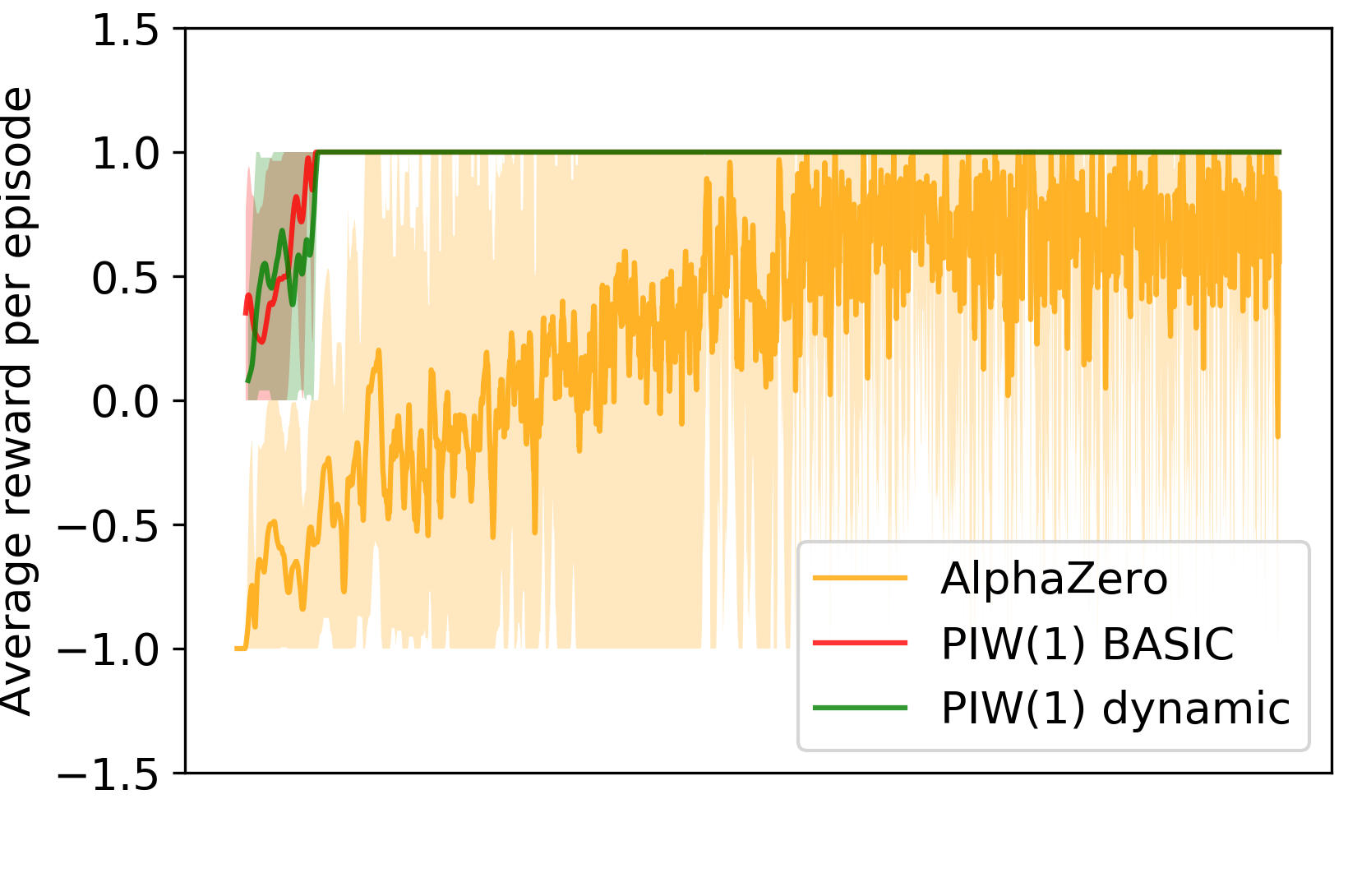}
\includegraphics[height=0.179\textheight]{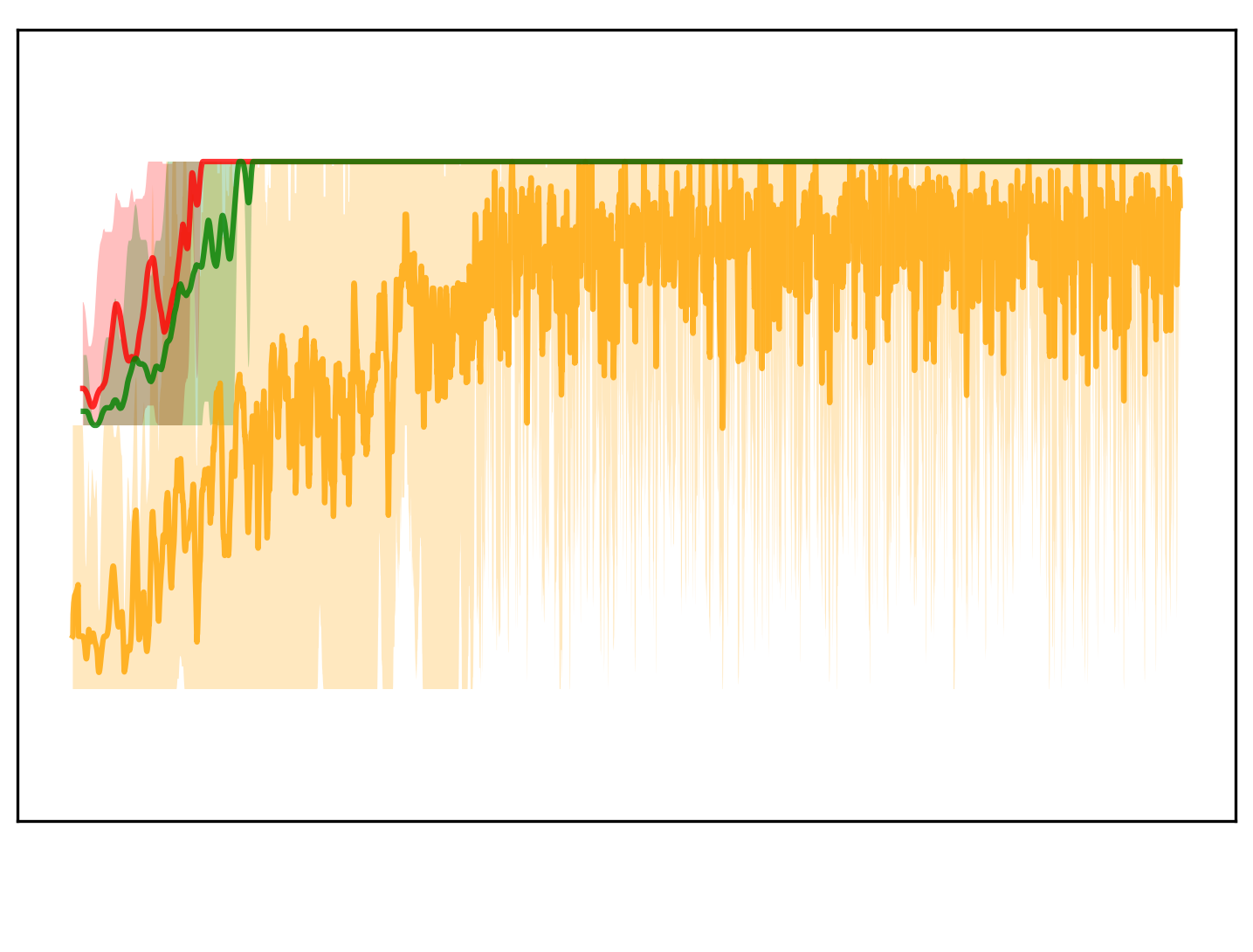}
\includegraphics[height=0.179\textheight]{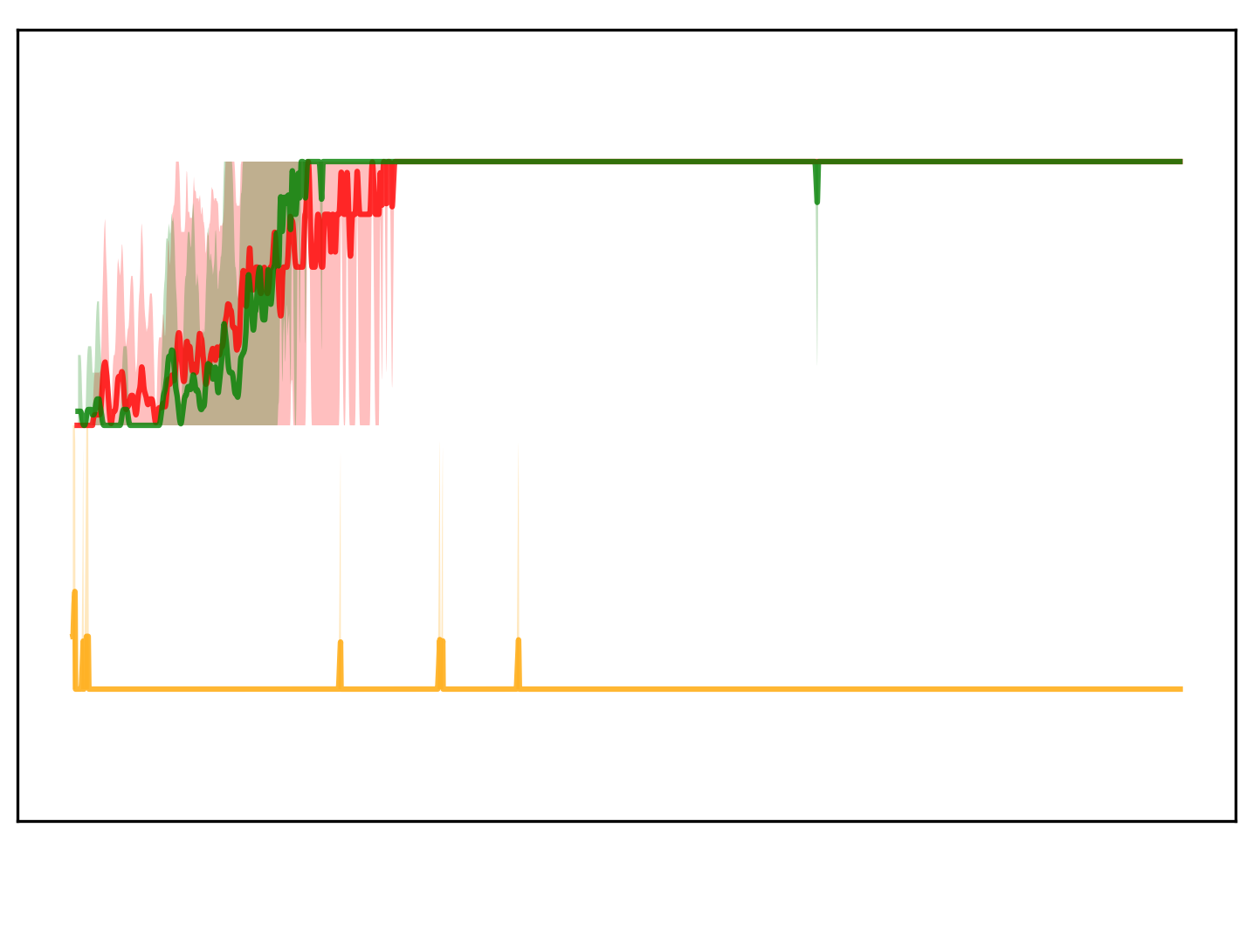}

\includegraphics[height=0.179\textheight]{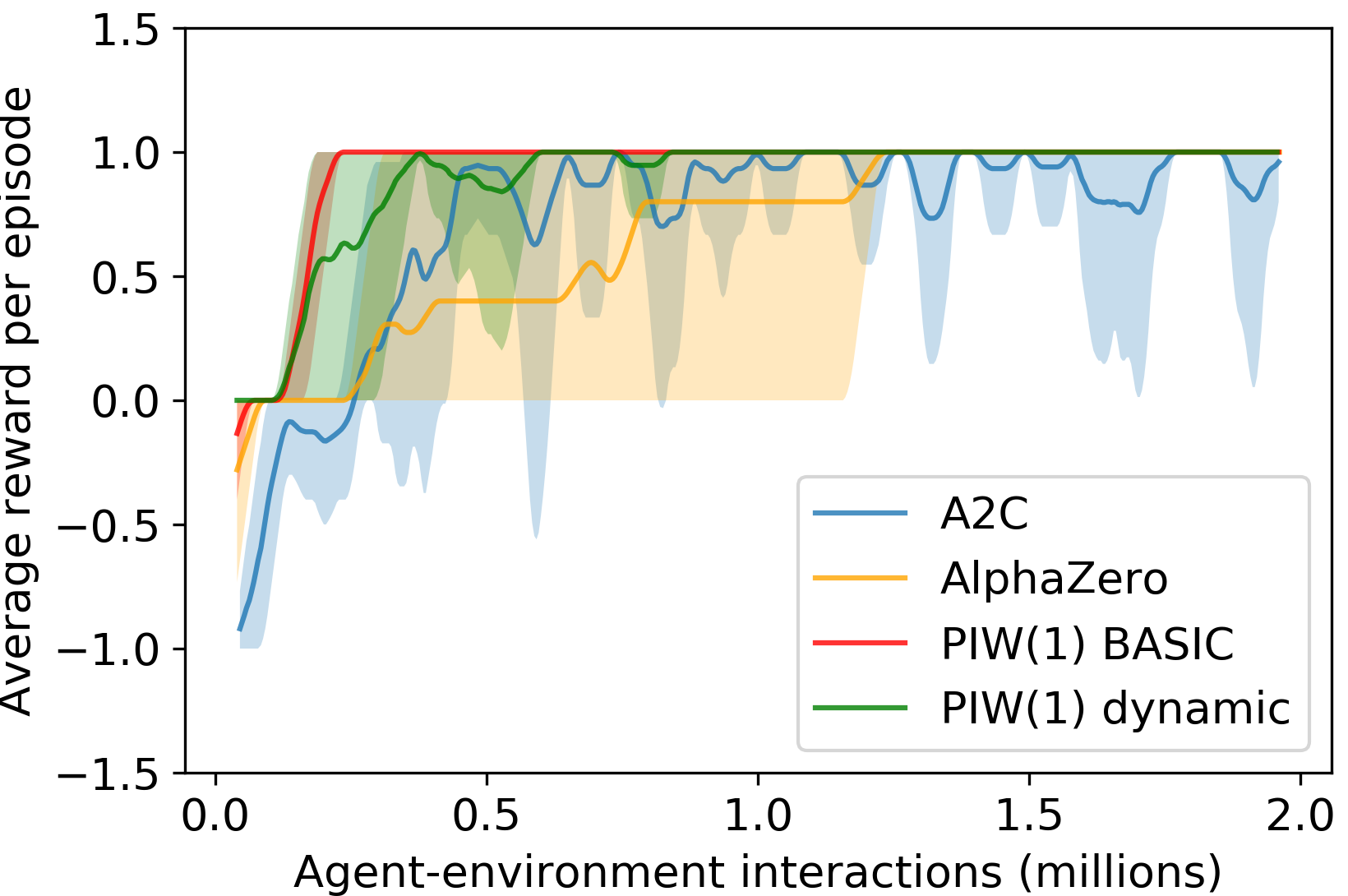}
\includegraphics[height=0.179\textheight]{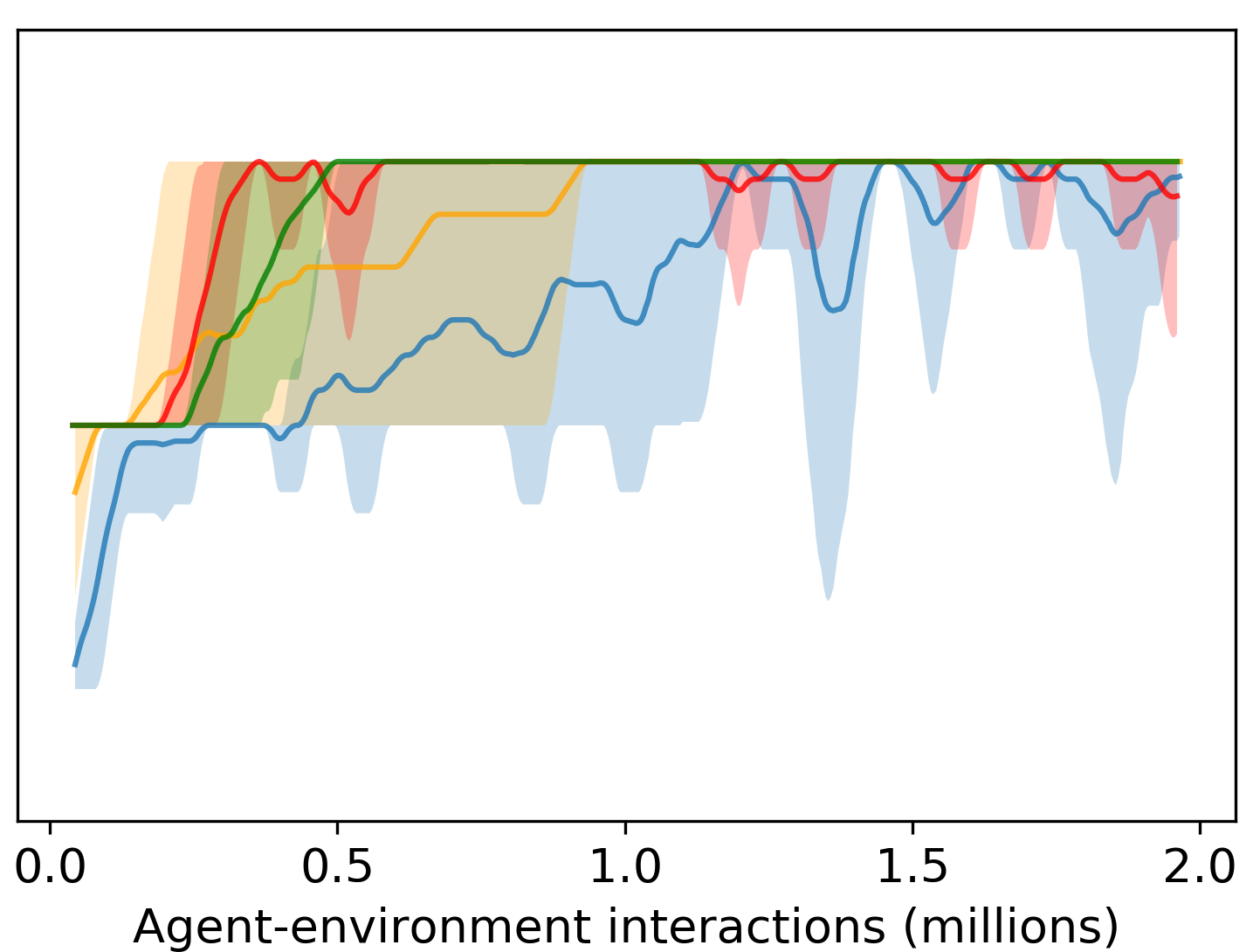}
\includegraphics[height=0.179\textheight]{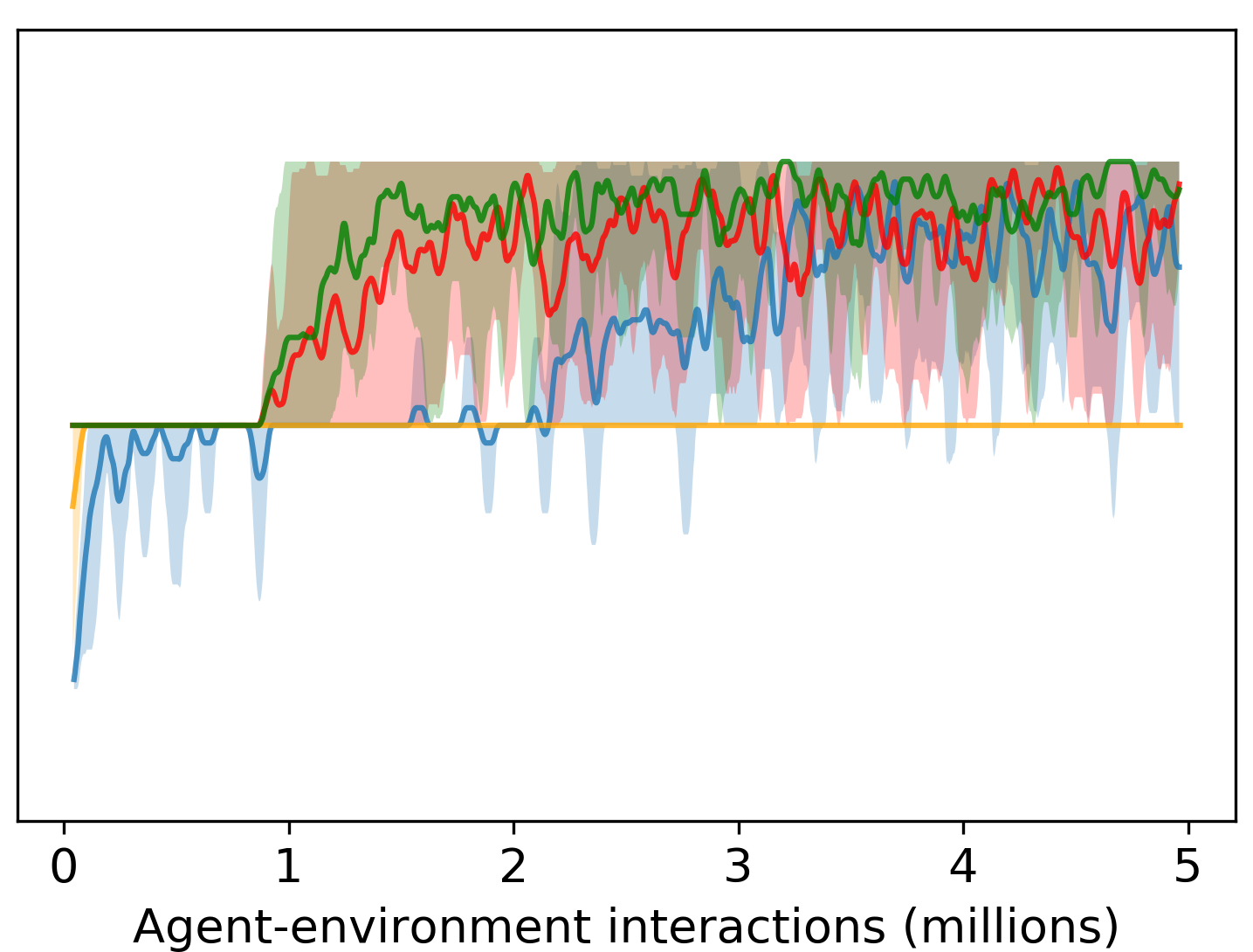}

\caption{Performance of PIW, AlphaZero and A2C in three simple mazes. Columns represent difficulty (1, 2 and 3 walls respectively), first row shows tree search-based algorithms lookahead performance and second row compares compact policies along learning. All plots are averages over five runs, and shades show the minimum and maximum score.}
\label{fig:simple-performance}
\end{figure*}

Figure \ref{fig:simple-performance} (top row) shows results comparing PIW and AlphaZero for the three mazes.
We plot the average reward as a function of the number of interactions with the environment.
As expected, the number of interactions with the environment required to solve the problem increases with the level of difficulty.
We observe that PIW outperforms AlphaZero in these environments using both dynamic and static features.
Surprisingly, the two variants of PIW show little difference, indicating that in this simple maze, the features can be learned easily.

The difference in performance between AlphaZero and the PIW variants is explained because AlphaZero needs to go through an optimal branch several times to increase its probability for action selection, since the policy estimate is based on counts. In contrast, PIW makes decisions solely on the rewards present on the tree. Thus, it may select a branch with low count after employing the budget to explore different parts of the state space. This, together with the use of rollouts that reach deeper states, makes IW more suitable for these challenging sparse-reward environments.


We also evaluate the learned policy of both algorithms every $20,000$ frames, and compare it with A2C. To do this, we 
choose a greedy policy $\widehat\pi_t$ with $\tau=0$, i.e. sampling uniformly between actions that present maximum probability. Figure \ref{fig:simple-performance} (bottom row) shows the results. We observe that PIW outperforms A2C in the three scenarios. This is explained again by means of the different types of exploration performed by the algorithms.
The exploration of A2C is purely random, and highly depends on the entropy loss factor. Since the reward in these environments is sparse, this factor needs to be high, which slows down the learning.

\subsection{Atari games}
We now consider the Atari benchmark. In this case, we only consider dynamic features. We set the maximum number of expanded nodes to $100$, a dataset of $T = 10^4$ transitions, L2-norm penalty of to $10^{-3}$, and linearly anneal the learning rate from $0.0007$ to $0.0005$. Just as in \citet{bandres2018planning}, we set the frame skip parameter to $15$ and all other hyperparameters equal to the previous experiments. In contrast to AlphaZero, PIW does not need to wait until an episode terminates to add transitions to the dataset. Thus, in these experiments transitions are added just right after they are generated, being directly available for the training step. Similar to previous work, the input for the NN consists of the last four grayscale frames, which are stacked together to form a 4-channel image.

\begin{table}
\caption{Comparison of accumulated reward of different width-based tree search methods. Performance of PIW (lookahead) is an average of 5 runs after 15M frames. Tree budget of either nodes or time shown in parentheses. Results from \citet{bandres2018planning}.}
\label{tab:atari-lookahead}
\begin{center}
\begin{small}
\begin{tabular}{lrrrr}
\toprule
 & IW & RIW & RIW & PIW \\
 & RAM & BPROST & BPROST & dynamic \\
Game & (1500) & (0.5s) & (32s) & (100) \\
\midrule
Breakout & 384.0 & 82.4 & 36.0 & 107.1 \\
Freeway & 31.0 & 2.8 & 12.6 & 28.65 \\
Pong & 21.0 & -7.4 & 17.6 & 20.7 \\
Qbert & 3,705.0 & 3,375.0 & 8,390.0 & 415,271.5 \\
\bottomrule
\end{tabular}
\end{small}
\end{center}
\vskip -0.1in
\end{table}

Table \ref{tab:atari-lookahead} shows results comparing PIW with previous width-based algorithms on the Atari games Pong, Freeway, Qbert and Breakout.
Our budget of 100 nodes at each tree expansion takes approximately 1 second in all games except Freeway, where the simulator is slower and takes 3 seconds.
In these four games, PIW clearly outperforms previous width-based algorithms based on pixel features, even compared to Rollout IW executions with a tree phase of 32 seconds.
Furthermore, our results are comparable to the ones achieved by~\citet{Lipovetzky2015ClassicalGames}, where the internal RAM state was used as the feature vector.
This suggests that using the policy is not only beneficial to guide the search, but also using its learned representation (our simple discretized features of the hidden layer) results in features that are exploited by IW. 
 Note that in our experiments we use a smaller budget of nodes (100 vs 1,500), which could explain the poorer performance in Breakout, for instance.



\begin{figure*}[!t]
\centering
\includegraphics[height=0.18\textheight]{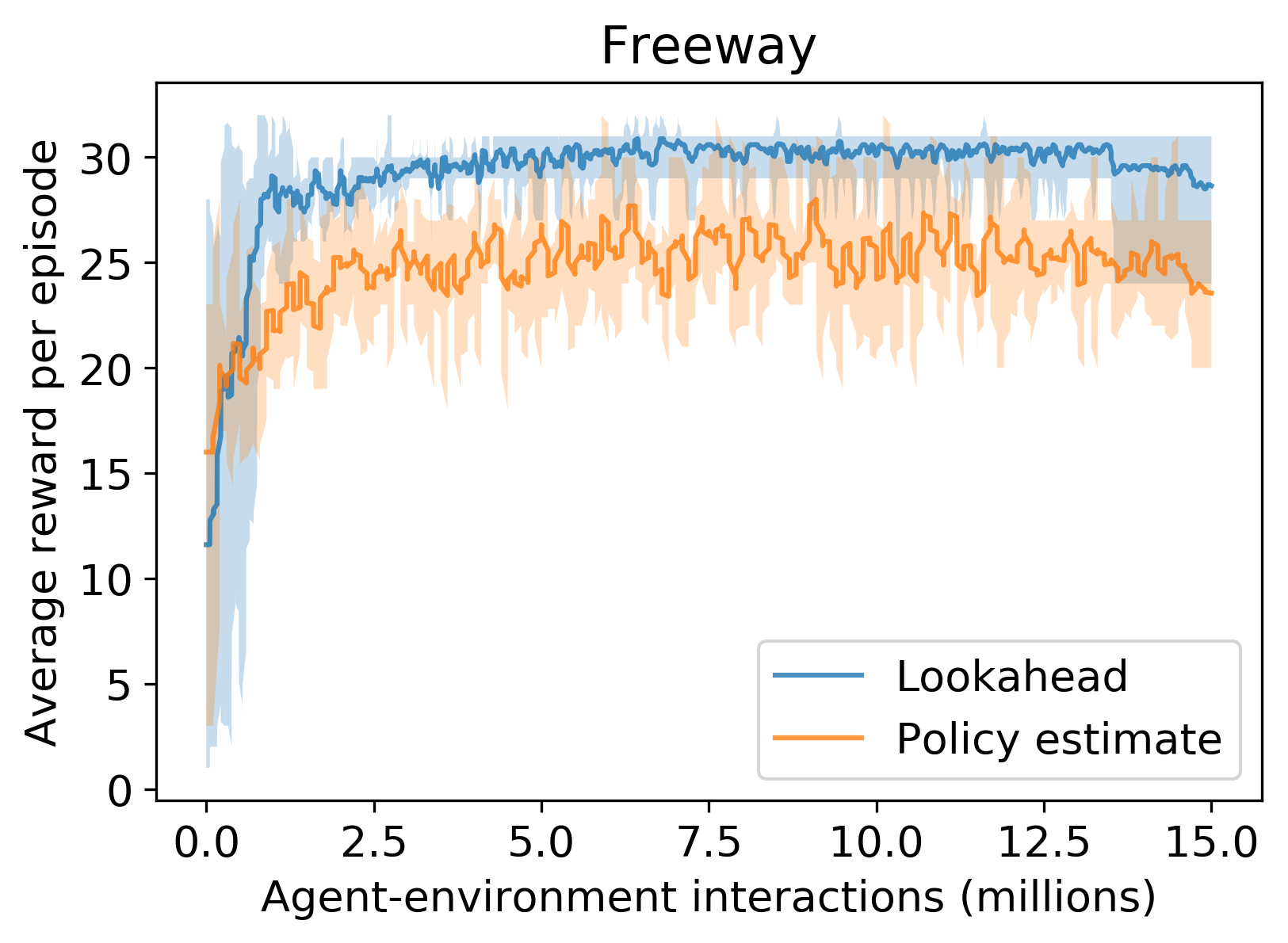}
\includegraphics[height=0.18\textheight]{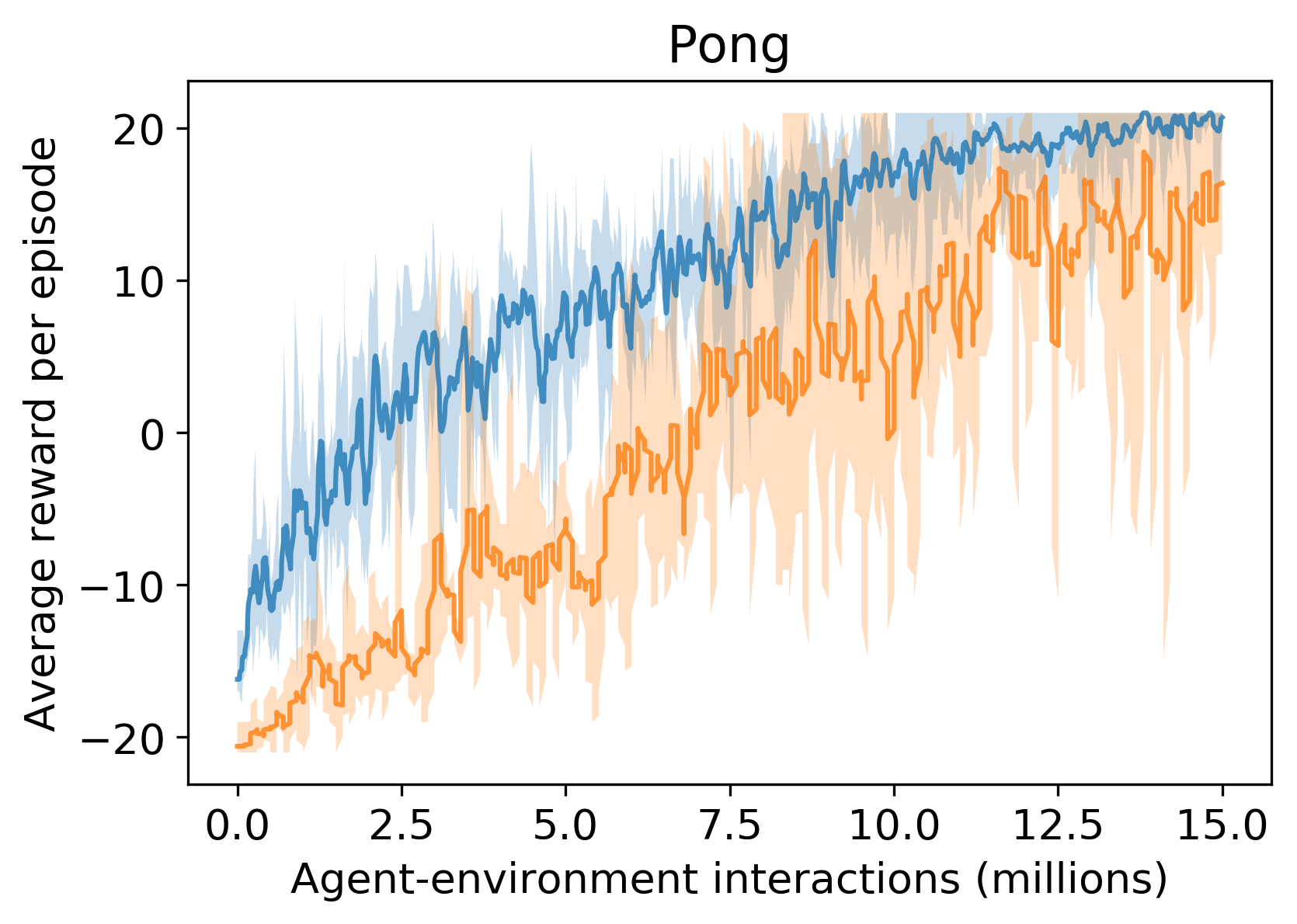}
\hspace{0.1\textwidth}
\includegraphics[height=0.18\textheight]{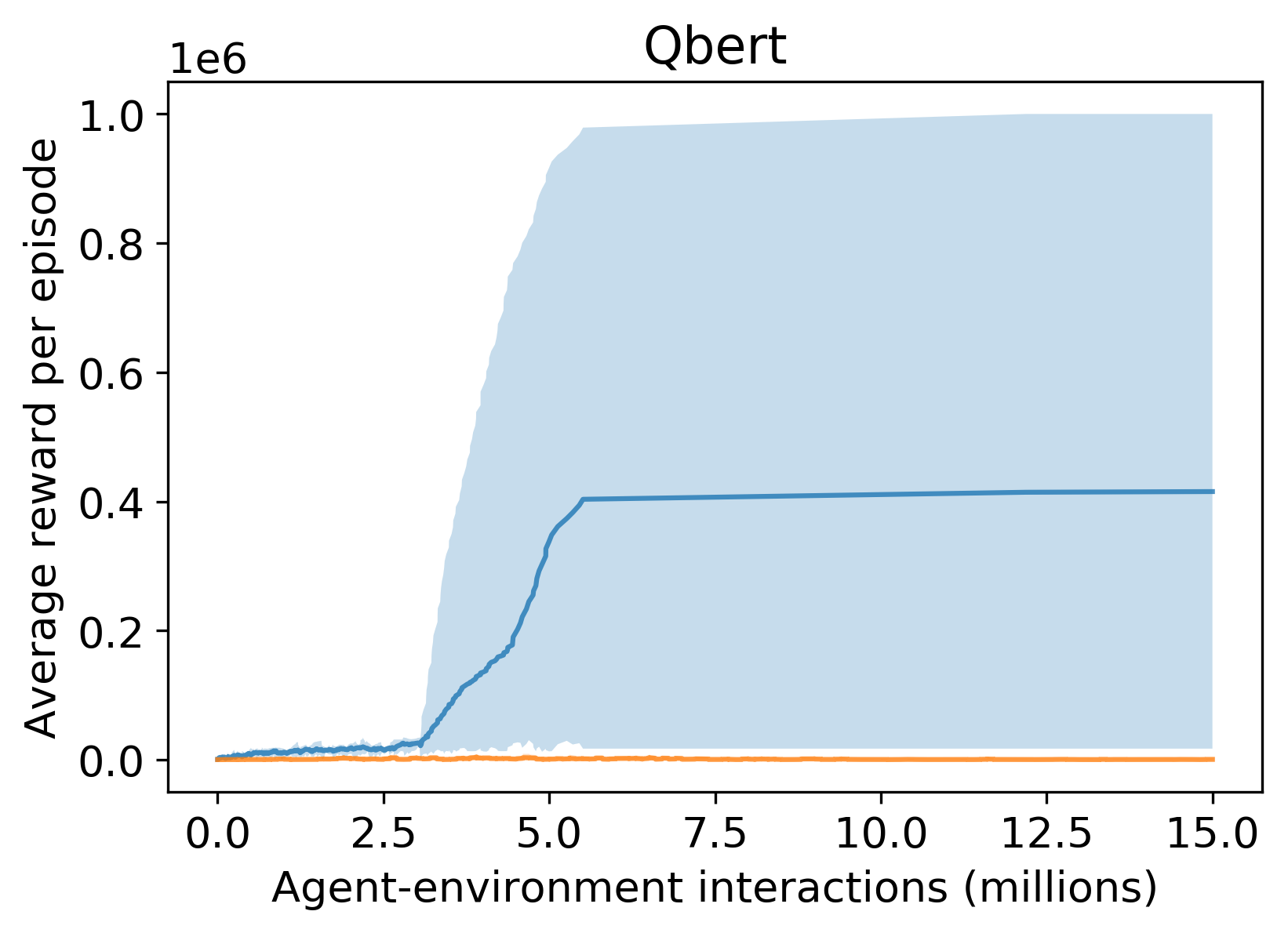}
\includegraphics[height=0.18\textheight]{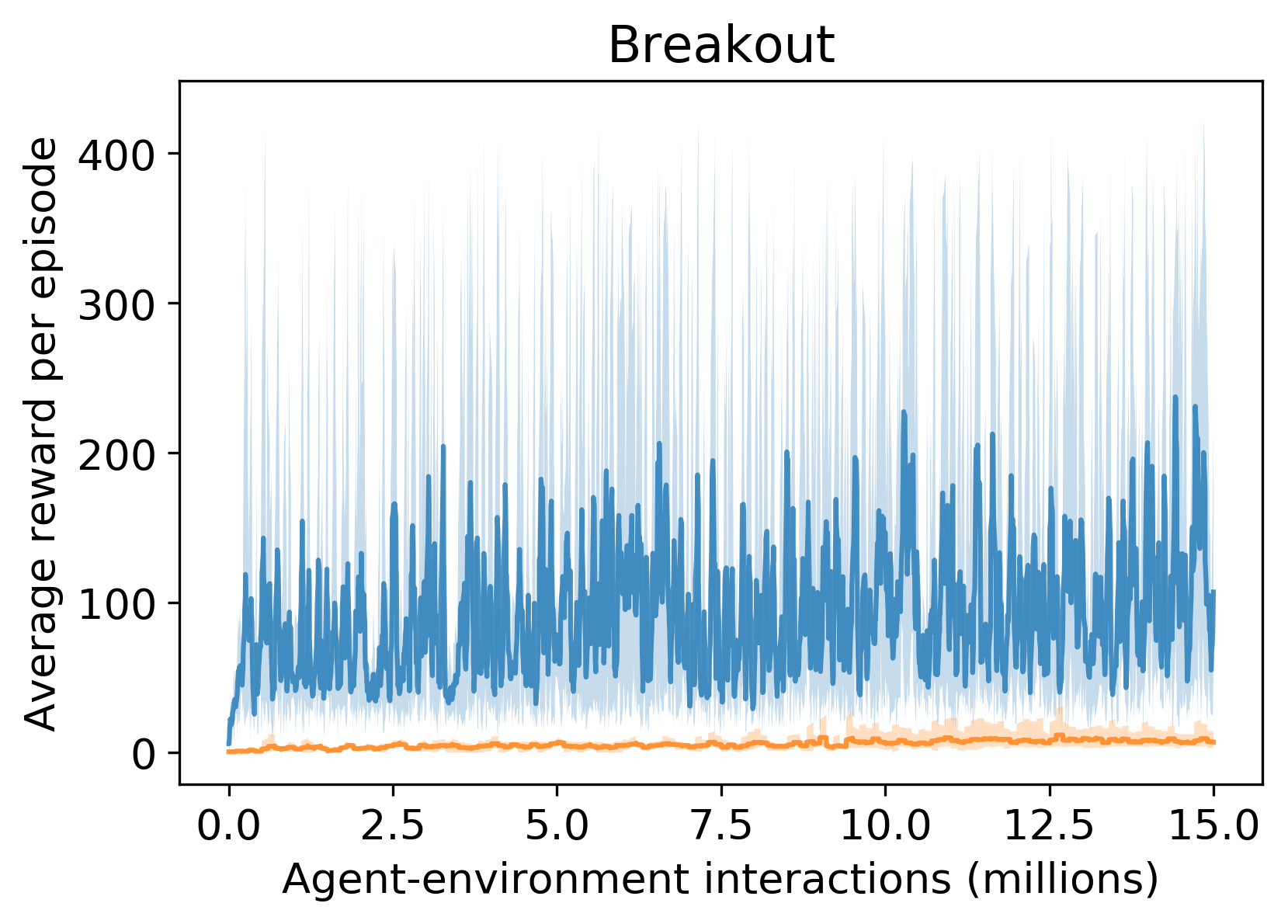}
\caption{Performance of PIW and its learned closed-loop controller in Freeway, Pong, Qbert and Breakout. Plots are averages of five runs and shades show the minimum and maximum score. Skipped frames are not counted as interactions.}
\label{fig:atari-performance}
\end{figure*}

In two executions of Qbert, the lookahead exploits a recently discovered glitch that leads to scores of near a million points \cite{chrabaszcz2018back}, while still achieving a remarkable score of around 30,000 in the other three.
Thus, the learned policy serves as a good heuristic for guiding the search. Nevertheless, the resulting policy itself is not a very good closed-loop controller.
This is shown in Figure \ref{fig:atari-performance}, where we show the performance of PIW (what we could call the \emph{teacher}) together with the learned closed-loop controller $\pi_\theta(\cdot|s)$, which is evaluated every $10^5$ interactions. In Freeway and Pong, the policy estimate is able to follow the target performance, although it does not match the results of the lookahead. In the game of Breakout we find a similar behavior as in Qbert, although the lookahead only improves in the beginning, resulting in a noisy performance.

\begin{table}[t]
\caption{Results for the policy learned by PIW compared to DQN, A3C and A3C+ (results taken from \citet{Bellemare2016UnifyingMotivation}).}
\label{tab:atari-policy}
\begin{center}
\begin{small}
\begin{tabular}{l@{\hspace*{5pt}}r@{\hspace*{5pt}}r@{\hspace*{5pt}}r@{\hspace*{5pt}}r@{\hspace*{5pt}}r}
\toprule
Game & Human & DQN & A3C & A3C+ & PIW \\
\midrule
Breakout & 31.8 & 259.40 & 432.42 & 473.93 & 6.9 \\
Freeway & 29.6 & 30.12 & 0.00 & 30.48 & 23.55 \\
Pong & 9.3 & 19.17 & 20.84 & 20.75 & 16.38 \\
Qbert & 13,455.0 & 7,094.91 & 19,175.72 & 19,257.55 & 570.5 \\
\bottomrule
\end{tabular}
\end{small}
\end{center}
\vskip -0.1in
\end{table}

Finally, we also compare the policy learned by PIW against some state-of-the-art RL methods.
Table \ref{tab:atari-policy} shows some preliminary results. 
Although the policy learned by PIW is not competitive, it is important to note that we use far less training samples (the horizontal axis include \emph{all} environment interactions, including the tree generation). Moreover, we used a frameskip of 15 based on previous work, instead of 4 as in DQN or A3C. This value may be correct for algorithms that perform a lookahead since all movements can be anticipated, but may be too high for estimating an action based solely on the current observation.

%% file: contents/conclusions.tex
\section{Conclusions}\label{sec:fin}
The exploration strategy of width-based planners is fundamentally different to existing RL methods, achieving state-of-the-art performance in planning problems and, more recently, in the Atari benchmark. In this work, we have brought width-based exploration closer to the RL setting.

Width-based algorithms require a factorization of states into features, which may not always be available. A second contribution of this paper is the use of the representation learned by the policy as feature space. We show how such a representation can be exploited by IW, achieving comparable results to using pre-defined features.

Our approach learns a compact policy using the exploration power of IW(1), which helps reaching distant high-reward states.
We use the transitions recorded by IW(1) to train a policy in the form of a neural network. Simultaneously, the search is informed by the current policy estimate, reinforcing promising paths.
Our algorithm operates in a similar manner to AlphaZero. It extends Rollout IW to use a policy estimate to guide the search, and interleaves learning and planning.
Differently from AlphaZero, exploration relies on the pruning mechanism of IW, it does not keep a value estimate, and the target policy is based on seen rewards rather than visitation counts.

Just like Monte-Carlo tree search, PIW requires access to a simulator. This is a departure from model-free RL, which uses the simulator as the environment. In this sense, we make use of the simulator to generate experience, and use that experience to learn a policy that can be used efficiently at execution time. We remark that Rollout IW (and consequently PIW) does not require storing and retrieving \emph{arbitrary} states, since rollouts always follow a trajectory and backtrack to the root state prior to the next rollout.


We have shown experimentally that our proposed PIW algorithm has superior performance in simple environments compared to existing RL algorithms. Moreover, we have provided results for a subset of the Atari 2600 games, in which PIW outperforms other width-based planner algorithms.
We have also evaluated the learned policy, and although it serves as a good heuristic to generate the tree, it fails to achieve the target performance of the lookahead. This could be due to several reasons (e.g. the frameskip may be too high, compared to what is used in DQN or A3C), and we leave for future work the necessary improvements to make the policy estimate match the lookahead performance. We would also like to investigate the use of a value estimate in our algorithm or to decouple learning features for IW from the policy estimate.